\documentclass[conference]{IEEEtran}
\usepackage{xcolor}
\usepackage{graphicx}
\usepackage{booktabs}
\usepackage{comment}
\usepackage[acronym]{glossaries}
\usepackage{algorithm}
\usepackage{algpseudocode}
\usepackage{caption}
\usepackage{subcaption}
\usepackage{lipsum}
\usepackage{float}
\usepackage{amsmath}
\makeglossaries
\newacronym{RPA}{RPA}{Robotic process automation}
\newacronym{LLM}{LLM}{Large Language Model}
\newacronym{ATS}{ATS}{Applicant Tracking System}
\newacronym{BPO}{BPO}{Business process outsourcing}
\newacronym{CV}{CV}{Curriculum vitae}
\newacronym{HR}{HR}{Human Resources}
\newacronym{PDF}{PDF}{Portable Document Format}

\begin{document}
\title{MLAR: Multi-layer Large Language Model-based Robotic Process Automation Applicant Tracking}
\author{
    Mohamed T. Younes * \\
    \textit{Computer Science Dept.}\\
    \textit{MSA University}\\
    Giza, Egypt\\
    mohamed.tarek61@msa.edu.eg
  \and
    Omar Walid\\
    \textit{Computer Science Dept.}\\
    \textit{MSA University}\\
    Giza, Egypt\\
    omar.walid2@msa.edu.eg
    \and
    Mai Hassan\\
    \textit{Computer Science Dept.}\\
    \textit{MSA University}\\
    Giza, Egypt\\
    maisalem@msa.edu.eg
    \and
    Ali Hamdi\\
    \textit{Computer Science Dept.}\\
    \textit{MSA University}\\
    Giza, Egypt\\
    ahamdi@msa.edu.eg
}

\maketitle

\begin{abstract}
This paper introduces an innovative \Gls{ATS} enhanced by a novel \Gls{RPA} framework or as further referred to as MLAR. Traditional recruitment processes often encounter bottlenecks in resume screening and candidate shortlisting due to time and resource constraints. MLAR addresses these challenges employing \Glspl{LLM} in three distinct layers: extracting key characteristics from job postings in the first layer, parsing applicant resume to identify education, experience, skills in the second layer, and similarity matching in the third layer. These features are then matched through advanced semantic algorithms to identify the best candidates efficiently. Our approach integrates seamlessly into existing \Gls{RPA} pipelines, automating resume parsing, job matching, and candidate notifications. Extensive performance benchmarking shows that MLAR outperforms the leading \Gls{RPA} platforms, including UiPath and Automation Anywhere, in high-volume resume-processing tasks. When processing 2,400 resumes, MLAR achieved an average processing time of 5.4 seconds per resume, reducing processing time by approximately 16.9\% compared to Automation Anywhere and 17.1\% compared to UiPath. These results highlight the potential of MLAR to transform recruitment workflows by providing an efficient, accurate, and scalable solution tailored to modern hiring needs.
\end{abstract}

\begin{IEEEkeywords}
Applicant Tracking System (ATS), Robotic Process Automation (RPA), Large Language Models (LLMs).
\end{IEEEkeywords}

\section{Introduction}

Recruitment is a vital yet challenging process, as HR teams face difficulties managing large volumes of applications. Manual candidate selection is labor intensive, error prone and involves risks overlooking qualified applicants or introducing biases. The integration of \gls{RPA} technologies augmented with \glspl{LLM} offers a solution to optimize recruitment workflows \cite{ATSimpact, RPAinHR}. Research shows that \gls{RPA} reduces time spent on repetitive administrative tasks, enabling HR professionals to prioritize strategic initiatives \cite{RPArecruitment, HRAutomationSurvey2023}.

Traditional \Gls{ATS} have played a pivotal role in automating aspects of the recruitment process, such as filtering resumes based on keyword matching \cite{RPAforresumeprocessing}. However, these systems face significant limitations. Keyword-based filtering often fails to capture the context of job requirements and applicant qualifications, resulting in suboptimal matching \cite{BertPaper}. Moreover, existing NLP-based text similarity and matching methods, while promising, struggle to process complex structures and large-scale recruitment data effectively and efficiently \cite{PospectCVpaper}. Research on ATS and RPA highlights the importance of semantic understanding in overcoming such challenges, with AI-powered models being increasingly adopted to enhance matching accuracy \cite{AiwithRPA, CombiningrpaandAI, AutomatedApplicantRanking}.

The emergence of Large Language Models (LLMs) has transformed text comprehension through advanced, context-sensitive processing. In recruitment, LLMs analyze unstructured text to identify essential attributes like education, experience, skills, and competencies, enhancing semantic interpretation of resumes and job descriptions \cite{ImprovingrecrruitmentprocessusingRPAandAI}. Innovative frameworks such as MockLLM replicate mock interviews to optimize candidate assessment, strengthening job-applicant alignment via refined semantic analysis \cite{LLMwithonlinejobseeking}. RPA systems integrated with AI and LLMs boost operational efficiency while minimizing biases and fostering ethical hiring practices \cite{ImprovingrecrruitmentprocessusingRPAandAI, ExplainableRPAHR}. Recent research demonstrates that specialized LLM-driven pipelines detect and mitigate biases in recruitment workflows, promoting equitable candidate selection \cite{LLMATS2024}.

This paper introduces \textbf{MLAR}, a novel Applicant Tracking System that leverages RPA and LLM technologies to automate end-to-end recruitment workflows. The system employs an LLM to identify critical attributes from resumes and job postings, compute compatibility metrics, and automatically notify top candidates. Unlike generic RPA tools, MLAR is specifically designed for high-volume resume analysis and recruitment enhancement, addressing shortcomings of current systems \cite{RPAuses, ResumeDataset}. Research shows that integrating LLMs into platforms like MLAR enhances semantic analysis and accelerates hiring workflows up to 11-fold compared to manual processes \cite{LLMinrecruitment}. Additionally, advanced semantic models, including GPT-based architectures, exhibit measurable F1-score improvements in aligning extensive resumes with job profiles \cite{SemanticMatchingGPT}, emphasizing the efficacy of LLM-centric methods.

The main contributions of this paper include: \begin{enumerate} \item An innovative ATS architecture combining \Glspl{LLM} for feature extraction and candidate-job alignment. \item Comprehensive evaluation of MLAR against top \Gls{RPA} platforms like UiPath and Automation Anywhere, demonstrating greater speed and efficiency \cite{UiPath, AutomationAnywhere}. \item Validation of the system’s capability to handle $2,400$ resumes within $3.5$ hours, averaging $5.4$ seconds per resume, surpassing existing solutions.\item An in-depth examination of \glspl{LLM} impact on recruitment automation, emphasizing semantic matching’s superiority over conventional keyword-based methods.\end{enumerate}

The remainder of this paper is organized as follows: Section~\ref{relatedwork} reviews related work in ATS, NLP techniques for recruitment, and RPA applications. Section~\ref{methodology} presents the problem formulation and challenges addressed by MLAR. Section~\ref{experiments} details the proposed system architecture and methodology. Section~\ref{results} discusses the experimental results and benchmarks. Finally, Section~\ref{conclusion} concludes the paper with insights and future directions.

\section{Related Work}\label{relatedwork}

Robotic Process Automation (RPA) is a technology designed to automate repetitive rules-based tasks traditionally performed by humans. RPA bots simulate human interactions with software systems, handling tasks such as data entry, email management, and workflow orchestration. Unlike traditional automation methods that require integration at the code level, RPA operates at the interface level, offering flexibility and ease of deployment across existing systems \cite{RPAadvantages}. Recent studies have shown how RPA can reduce the workload in recruitment processes and make them much faster \cite{RPArecruitment}.

Popular RPA platforms, such as UiPath \cite{UiPath} and Automation Anywhere \cite{AutomationAnywhere}, provide comprehensive tools for task automation across various industries, including recruitment. UiPath emphasizes user-friendly interfaces, featuring drag-and-drop functionality for building automation workflows. Its "Studio" environment allows developers and non-developers to create bots efficiently, supporting both attended and unattended bots. In contrast, Automation Anywhere offers cloud-native solutions with a focus on analytics. Its "Bot Insight" tool provides real-time tracking of bot performance, making it suitable for large-scale, data-driven operations. However, as pointed out by \cite{AiwithRPA}, while these platforms are powerful, they still need significant customization to be useful for recruitment tasks like parsing resumes or matching candidates to jobs.

Several research efforts have explored the application of RPA in recruitment. For example, \cite{RPAforresumeprocessing} demonstrated how RPA can automate resume parsing and categorization using document classification techniques. This approach utilized Naïve Bayes classifiers and custom-trained Named Entity Recognition (NER) models to extract essential details, including skills, experience, and education. But their system struggled with understanding the deeper context needed to match candidates to the right jobs. To tackle this, newer approaches have started combining RPA with AI, as discussed by \cite{CombiningrpaandAI}. AI tools add smarter decision-making abilities and improve how scalable these systems can be.

In more recent work, Large Language Models (LLMs) have started to play a big role in making Applicant Tracking Systems (ATS) smarter. For instance, \cite{BertPaper}, the authors proposed a BERT-based framework for evaluating resumes and job descriptions, calculating similarity scores to predict candidate suitability. Another example is the ProspectCV system introduced in \cite{PospectCVpaper}. It uses Gemini Pro, an advanced LLM, to process resumes and job descriptions, providing detailed compatibility scores and feedback for candidates. These systems are much better at understanding the meaning behind words, which makes them more accurate than traditional RPA systems. Research by \cite{LLMinrecruitment} also showed how LLMs could summarize resumes and grade candidates effectively, which saves time for recruiters.

Some studies even explored using LLMs for tasks like mock interviews. For example, \cite{LLMwithonlinejobseeking} suggested a system where an LLM acts as both the interviewer and the candidate to simulate job interviews. This kind of setup helps create a better understanding of whether a candidate is a good fit for the job. While these ideas are promising, they also raise important ethical questions about privacy and fairness in recruitment, as pointed out by \cite{ImprovingrecrruitmentprocessusingRPAandAI}. They highlight the need to balance innovation with ethical practices.

Even though these systems are improving, many of them still struggle with scalability and efficiency, especially when it comes to handling end-to-end recruitment workflows. To address these challenges, our proposed MLAR combines the automation of RPA with the intelligence of LLMs to deliver a state-of-the-art alternative to existing platforms. By using LLMs to extract important features from resumes and job postings, MLAR aims to make candidate-job matching faster and more accurate. This approach is supported by findings from \cite{AiwithRPA} and \cite{ImprovingrecrruitmentprocessusingRPAandAI}, who stress the importance of building systems that are both efficient and ethical.

This study positions MLAR as a specialized solution for recruitment, addressing the gaps in current systems and providing a more efficient, accurate, and scalable alternative to traditional RPA-powered ATS platforms. It takes advantage of the latest developments in RPA and LLM technologies to create a better recruitment process \cite{RPArecruitment, LLMwithonlinejobseeking}.

\section{Research Methodology}\label{methodology}

\subsection {Problem Formulation}
Recruitment is one of the most crucial yet complex functions within any organization. It directly impacts a company's ability to achieve its strategic objectives by ensuring the right talent is hired for the right roles. However, the recruitment process often faces significant challenges that hinder efficiency, and accuracy and scalability.

One of the most pressing challenges is managing the large volume of job applications received for each position, modern recruitment processes frequently involve handling thousands of resumes for a single job posting. As companies scale their operations or open multiple roles simultaneously, this volume multiplies, creating bottlenecks in the screening and evaluation phases. Handling such large datasets manually is both time-intensive and impractical. \cite{ATSimpact, RPAforresumeprocessing}.

The reliance on traditional methods further worsens these challenges. Manual resume screening is a time-consuming process prone to human errors. Recruiters may overlook highly qualified candidates due to fatigue or inefficiencies in handling large datasets. Additionally, unconscious biases can influence decision-making, resulting in inequities in candidate selection. For example, factors such as gender, ethnicity, or the formatting of a resume can unintentionally affect hiring decisions, compromising fairness and diversity in recruitment outcomes\cite{RPAinHR} \cite{ATSimpact}. Studies have shown that poorly formatted resumes, despite containing relevant information, are often disregarded during manual screening \cite{ImprovingrecrruitmentprocessusingRPAandAI}.

To address these challenges, many organizations have adopted Applicant Tracking Systems \Gls{ATS}, which partially automate the recruitment process. However, these systems are not without their limitations. Most traditional \Gls{ATS} rely on keyword-matching algorithms to filter resumes. While this approach is a step forward, it often fails to capture the full context of a candidate’s qualifications \cite{ATSimpact}. A candidate with relevant skills and experience may be excluded simply because their resume does not contain specific keywords matching the job description. This shortcoming results in a narrower pool of candidates for consideration, potentially eliminating highly suitable applicants. Moreover, \Gls{ATS} often require HR professionals to manually review and refine the system’s outputs, which reintroduces inefficiencies and diminishes the potential time savings.

Another critical challenge lies in the end-to-end recruitment workflow, which involves multiple stages, including job posting, resume parsing, candidate shortlisting, and notification. In traditional systems, these stages often require a combination of manual intervention and semi-automated processes, leading to delays and inconsistencies \cite{LLMinrecruitment, RPAadvantages}. The inefficiency of these workflows directly impacts the time-to-hire, reducing an organization’s ability to secure top talent in competitive markets. For companies operating at scale, these delays can result in lost opportunities and reduced organizational productivity.
\cite{ImprovingrecrruitmentprocessusingRPAandAI} \cite{AiwithRPA}

\subsection{MLAR model}
\textbf{Job Description and Resume Parsing} The system parses job descriptions and resumes to extract key features. Let \( J \) represent the set of all job descriptions, and \( R \) represent the set of all resumes. The system detects key features as follows:

\begin{equation}
F_J(j) = \{ f_1, f_2, \dots, f_m \}, \quad F_R(r) = \{ f_1, f_2, \dots, f_n \}
\end{equation}

where \( F_J \) and \( F_R \) are functions mapping job descriptions and resumes to feature sets.

\textbf{Semantic Matching Using Gemini LLM} To calculate the similarity score \( S \) between a job description \( j \) and a resume \( r \), a language model \( L \) is employed. The formula is given by:

\begin{equation}
S(j, r) = L(F_J(j), F_R(r))
\end{equation}

where \( L \) computes the semantic similarity.

\textbf{Ranking and Filtering} Resumes are ranked based on similarity scores, where the ranking is performed as follows:

\begin{equation}
\text{Rank}(r) = \text{sort}(\{ S(j, r) \mid r \in R \}, \text{descending})
\end{equation}

The top \( 3 \) resumes are selected using:

\begin{equation}
R_{\text{selected}} = \{ r_1, r_2, r_3 \} \subseteq R
\end{equation}

\textbf{Automated Communication} For each selected resume \( r_i \), the system generates an automated response \( C \) using the formula:

\begin{equation}
C(r_i) = \text{GenerateResponse}(r_i, j)
\end{equation}

where \( \text{GenerateResponse} \) is the function responsible for personalized communication.

\textbf{Performance Metrics and Benchmarking} The performance of MLAR is compared to UiPath and Automation Anywhere. Let \( T_{\text{UiPath}} \), \( T_{\text{AutomationAnywhere}} \), and \( T_{\text{MLAR}} \) represent the processing times for the respective systems. The comparison is performed using:

\begin{equation}
\Delta T = T_{\text{Benchmark}} - T_{\text{MLAR}}
\end{equation}

where \( T_{\text{Benchmark}} \) is the processing time for UiPath and Automation Anywhere.

\textbf{Continuous Operation Loop} The system operates continuously by monitoring new inputs. The process is described as:

$\quad \text{While } t \in T, \text{ perform MLAR process on } \{ J, R \}$
\begin{algorithm}
\caption{The MLAR Algorithm}
\label{alg:MLAR}
\begin{algorithmic}[1]
\State \textbf{Initialize} monitoring of the job descriptions $J$ and resumes $R$ 
\While{True}
    \State \textbf{Check} for new job descriptions or resumes in $J$ and $R$
    \If{new job description $j \in J$ or resume $r \in R$ is detected}
        \State \textbf{Parse} job description $j$ and resume $r$ to extract feature sets $F_J(j) \text{ and } F_R(r)$
        \State \textbf{Compute} similarity score $S(j, r)$ using LLM $L$
        \State \textbf{Rank} resumes based on similarity scores $Rank(r)$
        \State \textbf{Select} top $3$ resumes
        \For{each resume $r_i \in R_{\text{selected}}$}
            \State \textbf{Generate} personalized response $C(r_i)$ for $j$
            \State \textbf{Notify} job seeker associated with $r_i$
        \EndFor
    \Else
        \State \textbf{Ignore} unrelated or invalid inputs
    \EndIf
    \State \textbf{Log} all operations and results to the database
\EndWhile
\end{algorithmic}
\end{algorithm}

\begin{figure*}[t] \centering \includegraphics[width=0.6\textwidth]{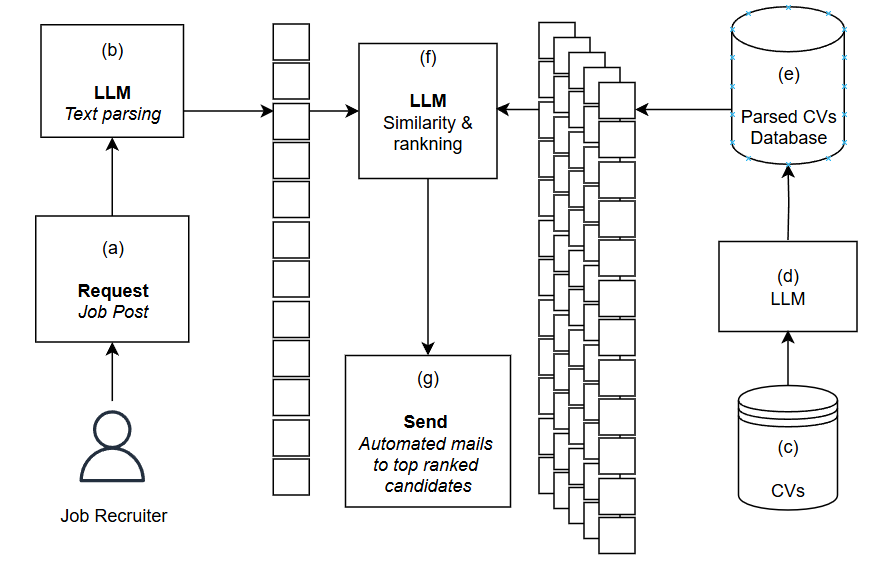} \caption{System architecture of the MLAR model. The diagram demonstrates the integration of Large Language Models (LLMs) at various stages of the recruitment process. It starts with job post requests from recruiters (a) and proceeds to LLM-based job description parsing (b). Candidate resumes (c) are parsed using LLMs (d) and stored in a parsed resume database (e). Finally, similarity and ranking are performed by the LLM (f), ensuring accurate matching between job descriptions and candidate profiles.} \label{fig:Sys_arch} \end{figure*}

\subsection{Tools and Dataset}
Firstly, the \Gls{LLM} - Gemini, serves as the backbone of the system, enabling advanced natural language processing tasks such as resume parsing and similarity matching. Its ability to extract structured information from unstructured documents like \Glspl{PDF} and job postings makes it an indispensable component of the system.

Secondly, RPA Tools: 
\begin{itemize}
    \item UiPath's automation capabilities were utilized to orchestrate the workflow, including file handling and email integration.
    \item Automation Anywhere: Similar to UiPath, Automation Anywhere was used to test automation scripts and compare performance metrics.
    \item MLAR: This is the custom-built \Gls{RPA} solution integrated with the Gemini API, which demonstrated superior efficiency and performance in processing resumes and job descriptions.
\end{itemize}

The Kaggle Resume Dataset was chosen for its diversity, containing 2400 resumes across 24 professions. These are: \Gls{HR}, Designer, Information-Technology, Teacher, Advocate, Business-Development, Healthcare, Fitness, Agriculture, \Gls{BPO}, Sales, Consultant, Digital-Media, Automobile, Chef, Finance, Apparel, Engineering, Accountant, Construction, Public-Relations, Banking, Arts, Aviation. All of the resumes were provided in \Gls{PDF} format. The dataset provided a realistic simulation of large-scale recruitment scenarios \cite{ResumeDataset}.

\section{Experimental Design} \label{experiments}

\begin{figure}[!h]    
\centering\includegraphics[scale=0.49]{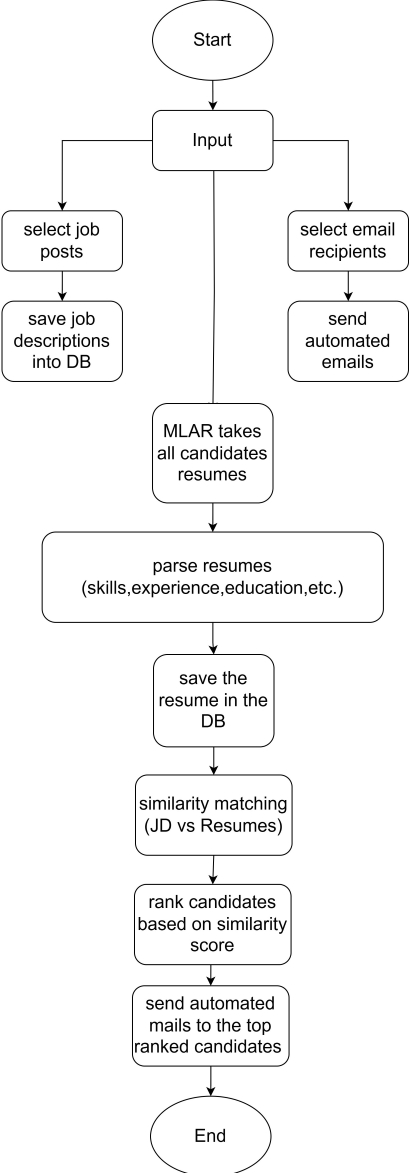}
    \caption{The MLAR flow diagram}
    \label{fig:MLAR_arch}
\end{figure}

The flow diagram in Figure \ref{fig:MLAR_arch} provides an overview of the MLAR system's operational workflow. The process begins with input from HR professionals, where job descriptions are saved into the database, and emails are prepared for notifications. Simultaneously, the system processes candidate resumes, parsing their information (e.g., skills, experience, and education) and saving the structured data into the database.

Next, the MLAR system performs a similarity matching between the job descriptions and resumes to compute scores based on alignment. The resumes are ranked accordingly, and the top-ranked candidates are identified. Finally, automated emails are sent to these candidates, streamlining the recruitment process and ensuring efficient communication.

The architecture of the MLAR system is centered around automating critical steps in the recruitment process. It optimizes efficiency by integrating \Gls{RPA} with AI-powered tools like \glspl{LLM}. The system architecture is divided into four main stages: Job Posting, Resume Parsing, Resume Matching, and Candidate Notification, which work cohesively to streamline the workflow.

Job Posting: This step (represented as part (a) in Figure \ref{fig:Sys_arch}) focuses on processing job descriptions provided by \Gls{HR} professionals. Initially, \Gls{HR} enters the recipient's email address for sending job postings. The system processes 24 job descriptions stored in a local folder, each representing a unique role across different departments. These job descriptions are then emailed to the specified address. To make the job descriptions machine-readable and ready for downstream tasks, the system leverages the Gemini \Gls{LLM} (part (b) in Figure 1), a state-of-the-art (SOTA) \Gls{LLM}. The \Gls{LLM} extracts crucial details from each job post, including Job Title, Required Skills, Experience Level, Educational Qualifications, and Additional Preferences if available.

Resume Parsing: Resume parsing (depicted as part (d) in Figure \ref{fig:Sys_arch}) is a key component of the system's architecture. The system processes 2400 resumes in PDF format (stored as part (c) in Figure \ref{fig:Sys_arch}), sourced from the Kaggle Resume Dataset, to simulate a high-volume recruitment scenario. Using Gemini \Gls{LLM}, the system extracts fields from each resume, such as candidate name, contact information (email and phone number), professional skills, work experience, educational background and predicted department classification. The parsed resumes are stored in a structured manner within the Parsed resumes Database (shown as part (e) in Figure \ref{fig:Sys_arch}). For example, if \Gls{LLM} predicts that a resume belongs to the "Engineering" category, it is stored in the "Engineering" department table within the database. This classification ensures that resumes are grouped logically, improving the speed and accuracy of subsequent processes.

Resume Matching: In this step (represented as part (f) in Figure \ref{fig:Sys_arch}), the system performs a similarity analysis between the parsed resumes and the job descriptions stored in MongoDB.  Gemini \Gls{LLM} computes a similarity score between each resume and job description, ranging from 0 to 100. This score quantifies the alignment of the candidate’s qualifications and experience with the job requirements. The results, including similarity scores and rankings, are stored in MongoDB. This approach allows the system to identify the most suitable candidates for each job in a data-driven manner, significantly reducing the time and effort required for manual screening.

Candidate Notification: Finally, the system automates candidate notifications (depicted as part (g) in Figure \ref{fig:Sys_arch}). After identifying the top three candidates for each job posting based on similarity scores, the system generates acceptance emails and sends them to the selected candidates. These emails include key details about the job and instructions for the next steps. This final step ensures that both job seekers and recruiters benefit from a seamless and automated experience, completing the overall workflow.

\section{Results and Discussion} \label{results}
The total time taken and time per resume for processing 2,400 resumes across UiPath, Automation Anywhere, and MLAR are summarized in Table 1:
\begin{table}[h!]
\centering
\begin{tabular}{@{}lcc@{}}
\toprule
\textbf{System} & \textbf{\begin{tabular}[c]{@{}c@{}}Total Time Taken\\(seconds)\end{tabular}} & \textbf{\begin{tabular}[c]{@{}c@{}}Time Per Resume\\ (seconds)\end{tabular}} \\
\midrule
\textbf{UiPath} & 15,258 & 6.45 \\
\textbf{Automation Anywhere} & 15,350 & 6.50 \\
\textbf{MLAR (Proposed system)} & 12,414 & 5.25 \\
\bottomrule
\textbf{}
\end{tabular}
\caption{Comparison of average automation speed between UiPath, Automation Anywhere, and MLAR for job posting, parsing, matching, and email-sending tasks.}
\label{tab:automation_speed}
\end{table}

The data show that MLAR was the most efficient system, completing all tasks (job posting, resume parsing, resume matching, and email notifications) in 12,414 seconds, averaging 5.25 seconds per resume. By comparison: UiPath required 15,258 seconds, or 6.45 seconds per resume, making it 22.8 \% slower than MLAR. Furthermore, Automation Anywhere took 15,350 seconds, or 6.50 seconds per resume, performing slightly worse than UiPath and 23.6 \% slower than MLAR.

Although the same Python scripts were used for all three systems, differences in execution speed can be attributed to how each platform manages external scripts and orchestrates processes, for example, UiPath is optimized for automation workflows; not running external Python scripts involves additional orchestration overhead, such as initializing environments, managing dependencies, and handling inter-process communication. This slightly increases execution time for compute-intensive tasks like resume parsing and matching. 
Moreover, Automation Anywhere introduces even greater latency, likely due to its reliance on cloud-based architecture for script execution. Although this makes it versatile for distributed workflows, it adds noticeable delays when processing large datasets such as 2,400 resumes. 

MLAR bypasses these orchestration layers by running scripts directly in Python, resulting in faster initialization and execution. This advantage is particularly evident in high-volume scenarios where the elimination of overhead translates to significant time savings.

The time differences may seem marginal per resume, but they become significant when scaled to large datasets, MLAR saves 2,844 seconds (47.4 minutes) compared to UiPath and 2,936 seconds (48.9 minutes) compared to Automation Anywhere. This time savings can be critical for real-world recruitment workflows, where speed directly impacts the ability to shortlist and contact candidates quickly, particularly in competitive hiring scenarios.

Although UiPath and Automation Anywhere are robust and versatile tools, their reliance on additional orchestration layers introduces inefficiencies when handling large-scale data processing. In contrast, MLAR’s direct execution model demonstrates superior performance, making it an optimal choice for automation scenarios that prioritize speed and scalability.

\subsection{Future Work}

The MLAR system achieved an accuracy of 63.45\% and a precision of 74.24\% in matching candidates with job requirements, laying a solid foundation for automated recruitment workflows. Although, the system demonstrates promising results, there is significant potential for improvement in future iterations.

The primary area for improvement involves the use of different fine-tuned \Glspl{LLM} specifically trained on recruitment datasets. By fine-tuning these local \Glspl{LLM} to better understand the relationships between job descriptions and resumes, the system can significantly improve its prediction accuracy. 

\section{Conclusion} \label{conclusion}
The integration of \gls{RPA} with advanced AI tools such as \glspl{LLM} has transformed recruitment automation, as demonstrated by the MLAR system. Our study shows that MLAR significantly improves efficiency, processing resumes in 5.25 seconds on average—faster than UiPath (6.45s) and Automation Anywhere (6.5s). This performance gain is due to MLAR’s seamless Gemini \Gls{LLM} integration and optimized data operations.

MLAR establishes a new benchmark for recruitment automation by reducing processing time and enhancing accuracy. Future work may focus on integrating additional datasets, refining \Gls{LLM} models, and incorporating features like interview scheduling and advanced applicant tracking. This study highlights the potential of AI-driven automation to optimize talent acquisition workflows efficiently.


%
\bibliographystyle{IEEEtran}

\end{document}